\documentclass[twoside,11pt]{article}

\usepackage{blindtext}

\usepackage[abbrvbib, preprint]{jmlr2e}

\usepackage[utf8]{inputenc} % allow utf-8 input
\usepackage[T1]{fontenc}    % use 8-bit T1 fonts
\usepackage{hyperref}       % hyperlinks
\usepackage{url}            % simple URL typesetting
\usepackage{booktabs}       % professional-quality tables
\usepackage{amsfonts}       % blackboard math symbols
\usepackage{amsmath}
\usepackage{amssymb}
\usepackage{nicefrac}       % compact symbols for 1/2, etc.
\usepackage{microtype}      % microtypography
\usepackage[dvipsnames]{xcolor}

\usepackage{graphicx}
\usepackage{comment}

\usepackage{lastpage}
\jmlrheading{23}{2022}{1-\pageref{LastPage}}{1/21; Revised 5/22}{9/22}{21-0000}{Yang et al}

\ShortHeadings{Fourier Neural Operator Downscaling}{Yang et al.}
\firstpageno{1}

\begin{document}

\title{Fourier Neural Operators for Arbitrary Resolution \\ Climate Data Downscaling}

\author{\name Qidong Yang \email qy707@nyu.edu\\
       \addr Mila Quebec AI Institute, Montreal, Canada\\
       \addr New York University, New York, USA
       \AND
       \name Alex Hernandez-Garcia\\
       \addr Mila Quebec AI Institute, Montreal, Canada\\
       \addr University of Montreal, Montreal, Canada
       \AND
       \name Paula Harder\\
       \addr Fraunhofer ITWM, Kaiserslautern, Germany\\
       Mila Quebec AI Institute, Montreal, Canada
       \AND
       \name Venkatesh Ramesh\\
       \addr Mila Quebec AI Institute, Montreal, Canada\\
       University of Montreal, Montreal, Canada
       \AND
       \name Prasanna Sattegeri\\
       \addr IBM Research,  New York, USA
       \AND
       \name Daniela Szwarcman\\
       \addr IBM Research, Brazil
       \AND
       \name Campbell D. Watson\\
       \addr IBM Research, New York, USA
       \AND
       \name David Rolnick\\
       \addr Mila Quebec AI Institute, Montreal, Canada\\
       McGill University, Montreal, Canada
}

\editor{My editor}

\maketitle

\begin{abstract}
Climate simulations are essential in guiding our understanding of climate change and responding to its effects. However, it is computationally expensive to resolve complex climate processes at high spatial resolution. As one way to speed up climate simulations, neural networks have been used to downscale climate variables from fast-running low-resolution simulations, but high-resolution training data are often unobtainable or scarce, greatly limiting accuracy. In this work, we propose a downscaling method based on the Fourier neural operator. It trains with data of a small upsampling factor and then can zero-shot downscale its input to arbitrary unseen high resolution. Evaluated both on ERA5 climate model data and on the Navier-Stokes equation solution data, our downscaling model significantly outperforms state-of-the-art convolutional and generative adversarial downscaling models, both in standard single-resolution downscaling and in zero-shot generalization to higher upsampling factors. 
Furthermore, we show that our method also outperforms state-of-the-art data-driven partial differential equation solvers on Navier-Stokes equations. Overall, our work bridges the gap between simulation of a physical process and interpolation of low-resolution output, showing that it is possible to combine both approaches and significantly improve upon each other.
\end{abstract}


\section{Introduction}
Climate simulations are running hundreds of years ahead to help us understand how climate changes in the future. Complex physical processes inside climate dynamical systems are captured by partial differential equations (PDEs), which are extremely expensive to solve numerically. As a result, running a long-term high-resolution climate simulation is still not feasible within the foreseeable future \citep{Balaji_2021}, even with the current fast-increasing computational power. Given neural networks' fast forward inference speed, deep learning has been applied to speed up climate simulations in the following two directions. 

First, neural networks are used as surrogate solvers to circumvent expensive numerical methods. More specifically, neural networks are trained with climate simulation data to approximate complex climate systems serving as climate model emulators. In recent years, neural network emulators have been successfully developed for modeling cloud, aerosol, and water systems \citep{Beucler_2019, Harder_2022_aerosol, Tran_2021}. Second, deep learning is also used to predict high-resolution versions of the lower-resolution outputs produced by climate simulators. Such a process is known as \textit{downscaling} in the climate science community and it resembles the problem of image super-resolution in the machine learning community. The recent works by \citet{H_hlein_2020, Price_2022, Groenke_2020} show that deep learning has achieved excellent performance at climate data downscaling on variables such as near-surface wind fields, precipitation, and temperature.  

%\citet{H_hlein_2020} used convolutional neural networks (CNNs) to downscale short-range forecasts of near-surface wind fields. A conditional generative adversarial network (GAN) was trained by \citet{Price_2022} via a custom training procedure and augmented loss function to downscale precipitation forecasts. \citet{Groenke_2020} proposed the first unsupervised statistical downscaling method based on normalizing flows to increase the resolution of temperature and precipitation data.

Limited by classic neural networks, which map between finite-dimensional spaces, neural network downscaling models typically have fixed input and output sizes. For a single trained model, it can only downscale input samples with a pre-defined upsampling factor. Inspired by the recent success of Fourier neural operator \citep[FNO]{li2021Fourier} for solving PDEs regardless of resolution, here we propose a novel FNO based zero-shot climate simulation data downscaling model, which is able to downscale input samples to arbitrary unseen high resolution by training only once on data of a low upsampling factor.

We evaluate our FNO downscaling model in three experiments: PDE integration, PDE solution downscaling and observational climate quantity downscaling. The PDE involved in the first two experiments is the Navier-Stokes equations, the central equation in most climate simulators, which describes physics status of a moving fluid (e.g., ocean or atmosphere). The observational climate quantity used in this work is the total column water content which we derived from the climate reanalysis data base ERA5 \citep{Hersbach_2020}. Climate downscaling models are generally applied to PDE based climate simulation as a post-processing tool to cheaply generate high-resolution simulation from a fast-running low-resolution numerical climate simulation model. Our FNO downscaling model fits this application well since smooth simulation data have a succinct representation in the Fourier basis, making it easier to be modeled by FNO with a truncated Fourier series. Evaluation on ERA5 water content data intends to examine to what extent our model can capture less smooth and noisy observational data.   

Downscaling experiments on Navier-Stokes solution data and water content data show that our model achieves great performance not only on the learned downscaling (i.e., the upsampling factor the model is trained on) but also on zero-shot downscaling (i.e., even higher upsampling factor unseen during training). The performance is even further improved when a softmax constraint layer \citep{Harder_2022} is stacked at the end of our model architecture to enforce conservation laws. In the PDE integration experiment, our model is used to downscale low-resolution solution from a numerical Navier-Stokes equation solver. The downscaled solution obtains significantly higher accuracy than that from an FNO equation solver---one of the state-of-the-art data-driven solvers \citep{li2021Fourier}. These results validate our model's potential to cheaply and accurately generate arbitrarily high-resolution climate simulation with fast-running low-resolution simulation as input.

\paragraph{Contributions} Our main contributions can be summarized as follows:
\begin{itemize}
\item To our best knowledge, we are the first to use FNOs for climate downscaling and to design an arbitrary-resolution downscaling model. 
\item Our FNO downscaling model performs significantly better than state-of-the-art deep learning-based downscaling models. %This performance gain gets more significant when constraint layers are applied to enforce conservation laws.
\item When trained on lower-resolution data and tested zero-shot on higher-resolution data, our method outperforms prior methods trained directly on higher-resolution data.
\item Combining our FNO downscaling model with a low-resolution physical solver, the resultant high-resolution solution outperforms that from a state-of-the-art data-driven solver.
\end{itemize}

\section{Related Work}
\subsection{Physics-Constrained Deep Learning for Climate System Emulation}
Due to their high approximation capacity and fast inference speed, neural networks have been widely applied for climate system emulation \citep{mccoy_2020, Watson_Parris_2021, Kasim_2022}. In such settings, it is essential for the output of a neural network not merely to be close to the ground truth, but also consistent with certain physical laws, which is important both for many downstream applications and for trustworthiness. Various works have attempted to embed physics constraints into neural network emulators by either adding violation penalty terms to the loss function (i.e., soft-constrained) or carefully designing a physics-preserving model structure (i.e., hard-constrained). \citet{Beucler_2021} applied soft-constrained and hard-constrained network emulators to atmospheric data. Their results showed that enforcing constraints, whether soft or hard, can systematically reduce model error, but the hard-constrained model is free of an accuracy-constraint trade-off. In addition, \citet{Daw_2020} developed constrained long short-term memory models to emulate lake water temperature dynamics. Their outcomes reflect the same pattern observed in \citet{Beucler_2021}.

\subsection{Deep Learning for Climate Downscaling}
Statistical downscaling of climate data using deep learning has attracted much attention over the last few years. Given the popularity of convolutional neural networks \citep[CNNs]{Dong_2015} and generative adversarial networks \citep[GANs]{Ian_2014} for super-resolution of natural images, they have become popular architecture choices for downscaling. \citet{chen_2022, Watson_2020, Chaudhuri_2020} used CNNs and GANs to downscale precipitation fields, while \citet{Harder_2023} used CNNs and GANs to downscale other quantities such as water content and temperature. So far, climate downscaling works have mainly focused on increasing the resolution in either spatial or temporal dimensions. Recently, \citet{Harder_2022} introduced a new spatiotemporal downscaling model (increasing resolution in both spatial and temporal dimensions), which stacks Deep Voxel Flow model \citep{Liu_2017} and ConvGRU network \citep{Ballas_2015}. It is able to generate accurate and reliable high-resolution outputs when a customized physics constraint layer is applied.

%Spatiotemporal downscaling (increasing resolution in both spatial and temporal dimensions), despite high practical value in many applications, has not been well studied. Some papers have looked at super-resolving multiple time steps at once, but not increasing the temporal resolution \citep{spatio_temp1, spatio_temp2}, whereas \citet{spatio-temp3} increases the temporal resolution by just treating time steps as different channels then using a standard super-resolution CNN for both spatial and channel dimensions. Recently, \citet{Harder_2022} introduced a new spatiotemporal downscaling model which stacks Deep Voxel Flow model \citep{Liu_2017} and ConvGRU network \citep{Ballas_2015}. It is able to generate accurate and reliable high-resolution outputs when a customized physics constraint layer is applied. 

\subsection{Fourier Neural Operators}
In a classic deep learning setting, a neural network is trained to approximate a function that forms a mapping between finite-dimensional spaces. Recent work by \citet{li2020neuop} generalized neural networks to neural operators, which can learn mappings between two infinite dimensional spaces (e.g., function spaces)---while keeping a finite set of parameters to define the neural architecture. %Neural operators can be constructed by a carefully designed network architecture with a set of finite parameters. 
They are typically trained in a supervised fashion to solve parameterized PDEs and demonstrate comparable performance to numerical solvers \citep{kovachki_2023}. Fourier Neural Operators (FNOs) \citep{li2021Fourier} extended neural operators to enable feature transformations with parameters defined in Fourier domain, resulting in an expressive and efficient architecture. FNOs became the first neural operator model to successfully learn a convergent solution operator for the Navier-Stokes equations in a turbulent regime.

\section{Methodology}
\subsection{Problem Setup}
Consider low-resolution input $\mathbf{a} \in \mathbb{R}^{d_a}$ and high-resolution output $\mathbf{b} \in \mathbb{R}^{d_b}$ with $d_a < d_b$. Traditional neural network downscaling models define a mapping $f: \mathbb{R}^{d_a} \rightarrow \mathbb{R}^{d_b}$ from low-resolution input $\mathbf{a}$ to high-resolution output $\mathbf{b}$. This formulation induces a limitation where the downscaled output resolution is fixed to be $d_b$. We propose the following formulation to relax this limitation to achieve arbitrary resolution downscaling.

Instead of looking for a mapping between two finite-dimensional spaces, our methodology learns a mapping from a finite-dimensional space to an infinite-dimensional space. Namely, this mapping takes in low-resolution input $\mathbf{a} \in \mathbb{R}^{d_a}$ and outputs a function $\mathbf{u}\in \mathcal{U}$ of which a high-resolution observation $\mathbf{b}$ is a discretization. We denote this mapping as: $G^{\dagger}: \mathbb{R}^{d_a} \rightarrow \mathcal{U}$, where $\mathcal{U} = \mathcal{U}(D; \mathbb{R}^{d_u})$ is a Banach space of functions taking values in $\mathbb{R}^{d_u}$ at each point from a bounded open set $D \subset \mathbb{R}^{d}$. $D$ can be viewed as a $d$-dimensional hypercube. As a result, arbitrarily high-resolution outputs can be obtained by evaluating $\mathbf{u}$ at arbitrarily many points from $D$.

Suppose we have observations $\{\mathbf{a}_{j}, \mathbf{u}_{j}\}_{j=1}^{N}$, where $ \mathbf{a}_{j}$ is an i.i.d. low-resolution sample and $\mathbf{u}_{j} = G^{\dagger}(\mathbf{a}_{j})$, function interpolating the high-resolution counterpart, is possibly corrupted with some random noise. We aim to construct a parametric map as follows to approximate $G^{\dagger}$:
\begin{equation}
    G: \mathbb{R}^{d_a} \times \Theta \rightarrow \mathcal{U} \; \; \; \; \; \; \text{or equivalently,} \; \; \; \; \; \;  
    G_{\theta}: \mathbb{R}^{d_a} \rightarrow \mathcal{U}, \; \theta \in \Theta,
\end{equation}
where $\Theta$ is a finite-dimensional parameter space. We aim to find a ${\theta}^{\dagger} \in \Theta$ such that $G(\mathbf{a}, {\theta}^{\dagger}) = G_{\theta^{\dagger}}(\mathbf{a})$ is close to $G^{\dagger}(\mathbf{a})$, which can be formulated as an optimization problem:
\begin{equation}
    \theta^{\dagger} = \arg \min_{\theta \in \Theta} \mathbb{E}_{\mathbf{a}} [C(G(\mathbf{a}, \theta), G^{\dagger}(\mathbf{a}))],
\end{equation}
where $C: \mathcal{U} \times \mathcal{U} \rightarrow \mathbb{R}$ is a cost functional measuring the distance in $\mathcal{U}$. In the following experiments, we take $\mathbf{a}_{j}$ as a single channel low resolution image and $\mathbf{u}_{j} \in \mathcal{U}((0, 1)^{2}; \mathbb{R})$ as a function interpolating its high resolution counterpart.

Note that our data $\mathbf{u}_{j}$ are functions. In practice, to work with $\mathbf{u}_{j}$ numerically, we assume access to point-wise evaluations of it, which is denoted as $u_{j}$. It is generated by a discretization operator $T$ applied to $\mathbf{u}_{j}$. Formally,
\begin{equation}
u_{j} = T(\mathbf{u}_{j}, \mathbb{D}) = \{\mathbf{u}_{j}(x_1), \dots, \mathbf{u}_{j}(x_n)\},
\end{equation}
where $\mathbb{D} = \{x_1, \dots, x_n \} \subset D $ is a $n$-point discretization of the domain $D$. 

\subsection{Implementation}
\label{sec:implementation}
Unlike neural operators which map between function spaces, the downscaling model $G_{\theta}$ defined in the previous section learns a mapping from a vector space to a function space. To achieve this transformation, a discretization inversion operator (mapping from a vector to a function) denoted as $T^{-1}$ is applied inside $G_{\theta}$. A neural network (mapping between vectors) denoted as $f_{\theta}$ and a neural operator (mapping between functions) denoted as $\mathcal{F}_{\theta}$ are also stacked before and after the discretization inversion  operator to increase the capacity of the whole model. Therefore, we construct $G_{\theta}$ as follows:
\begin{equation}
    G_{\theta}(\mathbf{a}) := \mathcal{F}_{\theta}(T^{-1}(f_{\theta}(\mathbf{a}))).
\end{equation}
$f_{\theta}: \mathbb{R}^{d_a} \rightarrow \mathbb{R}^{d}$ is a vector-valued function parameterized by a neural network; the discretization inversion operator $T^{-1}: \mathbb{R}^{d} \rightarrow \mathcal{E}(D; \mathbb{R}^{d_e})$ is implemented as an interpolation scheme, which interpolates the output of $f_{\theta}$ as a function $\mathbf{e} \in \mathcal{E}$ over domain $D$; and $\mathcal{F}_{\theta}: \mathcal{E} \rightarrow \mathcal{U}$ is a functional operator parameterized by a neural operator \citep{li2020neuop}. Here $T^{-1}$ can be a very simple interpolation scheme (e.g.,~linear interpolation) without hurting the expressiveness of the overall model $G_{\theta}$. There are two reasons for it. First, $f_{\theta}$ is able to learn a high-dimensional embedding such that a simple interpolation of it would retain high expressiveness for the target with lower dimensionality. Second, $\mathcal{F}_{\theta}$ can learn a highly non-linear operator to apply complicated transformations to the interpolated function $\mathbf{e} = T^{-1}(f_{\theta}(\mathbf{a}))$ despite of the simple components of $\mathcal{F}_{\theta}$ \citep{li2021Fourier}. During inference, by evaluating $\mathbf{e}$ at a specific resolution over a domain $D$, we can obtain the downscaled output at any desired resolution. 

\begin{figure}[ht]
\begin{center}
\includegraphics[width=\textwidth]{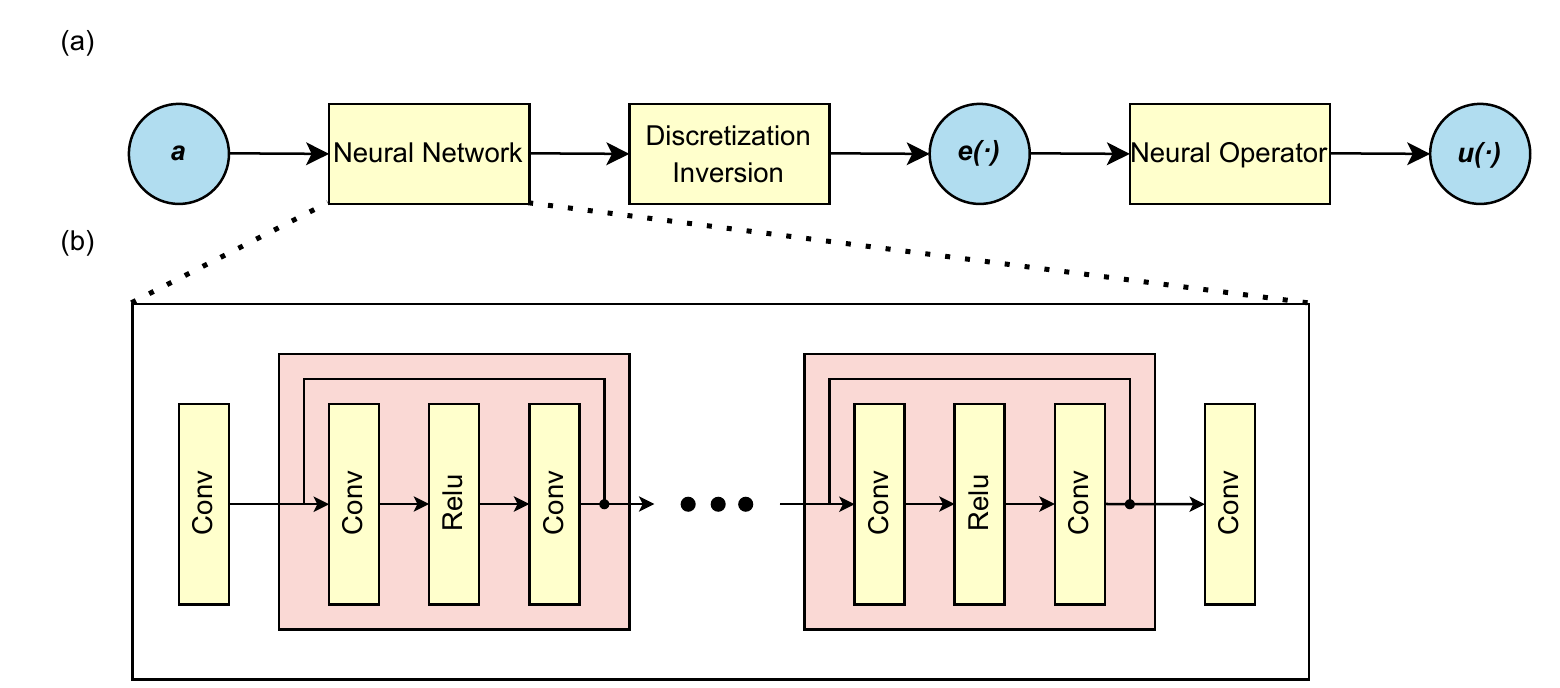}
\end{center}
\caption{The upper panel shows the overall structure of our Fourier neural operator downscaling model denoted by $G_{\theta}$. The low-resolution input $\mathbf{a}$ goes through a neural network $f_{\theta}$ and a discretization inversion operator $T^{-1}$. Then an embedding function $\mathbf{e}(\cdot)$ over domain $D$ is returned. Finally, a neural operator $\mathcal{F}_{\theta}$ takes in $\mathbf{e}(\cdot)$ and outputs the target function $\mathbf{u}(\cdot)$ which interpolates the high-resolution observation of $\mathbf{a}$. The lower panel shows the detailed architecture of $f_{\theta}$, which starts and ends with a convolutional layer, sandwiching a series of convolutional residual blocks.}
\label{fig:model_structure}
\end{figure}

In this work, $f_{\theta}$ is represented by a residual convolutional network inspired by the generator architecture of a widely used super-resolution GAN \citep{Wang_2018}; an FNO is implemented for $\mathcal{F}_{\theta}$; and bicubic interpolation is used as $T^{-1}$. Figure \ref{fig:model_structure}(a) shows an illustration of the overall structure of our proposed downscaling FNO (DFNO) model, denoted by $G_{\theta}$. The detailed architecture of neural network $f_{\theta}$ is pictured in Figure \ref{fig:model_structure}(b). For the FNO $\mathcal{F}_{\theta}$, we use the same architecture as described in \citet{li2021Fourier}.

\section{Experiments}
% TODO: add short intro paragraph summarising the section.
\subsection{Downscaling PDE Data}
\label{sec:PDE_downscaling}
In order to evaluate the performance of our model to downscale PDE data, we used a dataset solving the 2D Navier-Stokes equation for a viscous and incompressible fluid in vorticity form \citep[Section 5.3]{li2021Fourier}. The equation was numerically solved $10000$ times at resolution $64 \times 64$ with randomly sampled initial conditions. Each solution was integrated for $50$ time steps with a viscosity of $\text{10}^{-4}$. Out of $10000$ solutions, $7000$, $2000$, and $1000$ solutions were sampled as train, validation, and test sets, respectively. The solutions at each time step were then downsampled via average pooling to resolutions $32 \times 32$ and $16 \times 16$. Our PDE downscaling dataset consists of the solutions along with the downsampled versions. 

Following implementation details specified in Section \ref{sec:implementation}, we constructed our DFNO model and trained it on the PDE downscaling dataset with upsampling factor 2 (i.e., $16 \times 16 \rightarrow 32 \times 32$), and then evaluated it at both 2 times (learned) and 4 times (zero-shot) downscaling. 

As baselines for comparison, we trained two CNN (CNN-2 and CNN-4) and two GAN (GAN-2 and GAN-4) downscaling models with pre-defined upsampling factors 2 and 4. The network architectures follow the design in the paper by \citet{Harder_2022}. The baseline models were trained on datasets of their corresponding upsampling factors, and their outputs were then adjusted to achieve the desired resolution for evaluation. Downscaling outputs from 2 times models (CNN-2 and GAN-2) increase their resoultion to 4 times by model recursion and bicubic interpolation. Correspondingly, downscaling outputs from 4 times models (CNN-4 and GAN-4) decrease their resolution to 2 times by average pooling and bicubic interpolation. As an additional simple, non-deep learning baseline, we also considered bicubic interpolation \citep{de_Boor_1962} to the target resolution. 

For reliable usage of downscaled results in downstream tasks, it is important for results to be both close to the ground truth and physically consistent. \citet{Harder_2022} showed that a softmax constraint layer can effectively enforce conservation laws in neural networks for downscaling, without decreasing accuracy. Thus, we conducted another set of experiments where all aforementioned models include an additional softmax constraint layer at its end to generate physically consistent outputs.

\begin{figure}[htb]
    \centering
    \includegraphics[width=\textwidth]{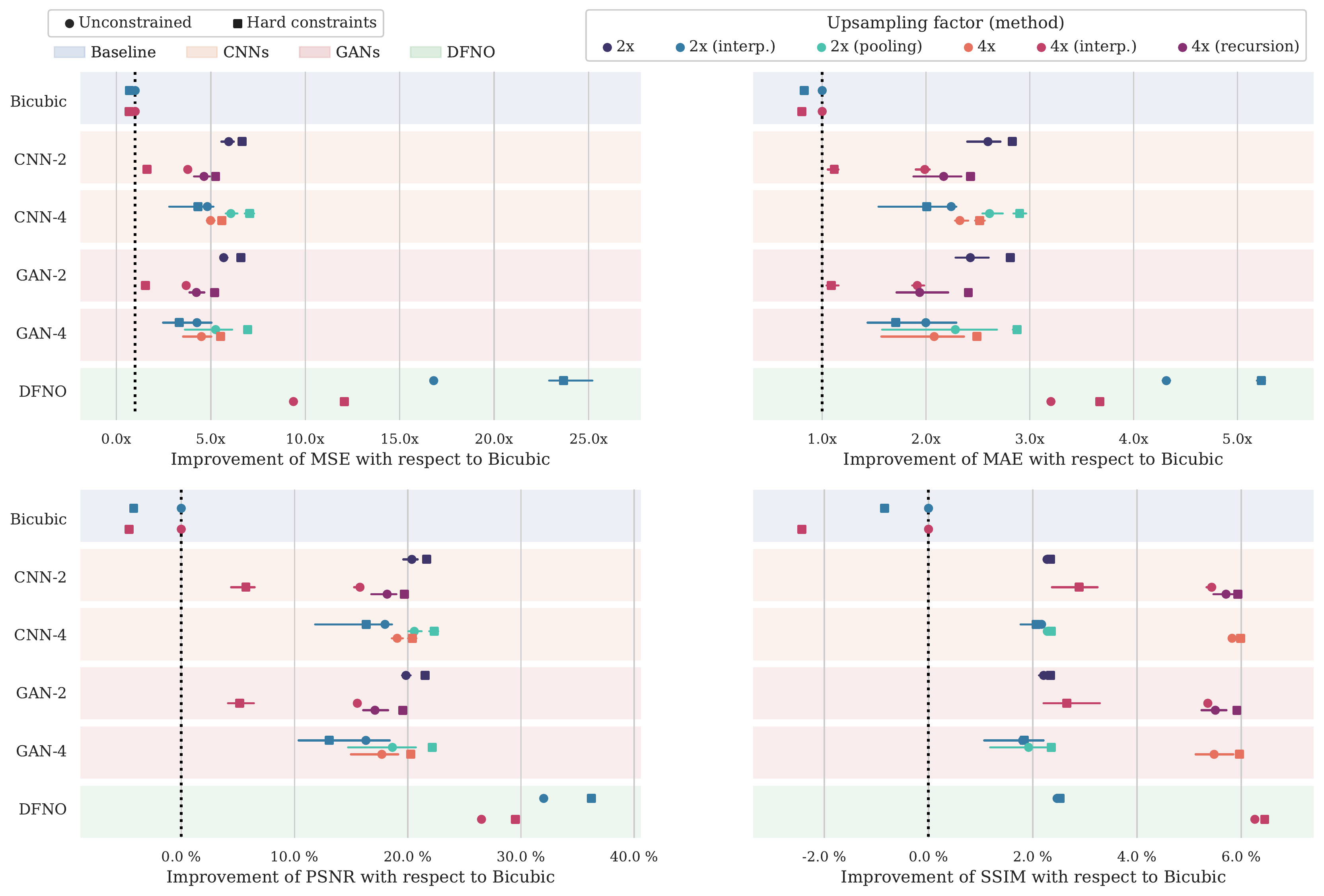}
    \caption{Metrics for downscaling models applied to the PDE dataset. Downscaling models CNN-2 (CNN-4) and GAN-2 (GAN-4) are trained with 2 times (4 times) downscaling data; the DFNO model is only trained with 2 times downscaling data. Each downscaling model is evaluated on both 2 times and 4 times downscaling. The 2 times downscaling outputs by CNN-2 and GAN-2 increase their resolution to 4 times through model recursion and bicubic interpolation. The 4 times downscaling outputs by CNN-4 and GAN-4 decrease their resolution to 2 times through average pooling and bicubic interpolation. Square (dot) denotes constrained (unconstrained) models. The metric mean and confidence interval from 3 runs are shown relatively to unconstrained bicubic interpolation. Model performance is evaluated by comparing marks of the same upsampling factor: cold colors for 2 times and warm colors for 4 times. Metric numerics and more details can be found in Tables \ref{tab:PDE_upsample_unconstrained_interp}, \ref{tab:PDE_upsample_unconstrained_other}, \ref{tab:PDE_upsample_constrained_interp}, and \ref{tab:PDE_upsample_constrained_other}.}
    \label{fig:PDE_upsample_metric}
\end{figure}

To evaluate each downscaling model, we computed the improvement with respect to the unconstrained bicubic baseline. In particular, in the case of error metrics, that is the mean squared error (MSE) and mean absolute error (MAE), the improvement was computed as the error of the bicubic baseline divided by the error of the evaluated model. In the case of the peak signal-to-noise ratio (PSNR) and the structural similarity index measure (SSIM), the improvement was computed as $100\times (M - B) / B$, where $B$ is the result of the bicubic baseline and $M$ is the result of the evaluated model. These derived relative results facilitate the comparison across models. These results are summarized visually in Figure~\ref{fig:PDE_upsample_metric} and we provide the evaluation numerical details in Tables \ref{tab:PDE_upsample_unconstrained_interp}, \ref{tab:PDE_upsample_unconstrained_other}, \ref{tab:PDE_upsample_constrained_interp}, and \ref{tab:PDE_upsample_constrained_other}. 

%To our surprise, DFNO performs worse on reconstructing inputs than downscaling inputs. However, it is not a major issue because downscaling models are intended to increase input resolution. 

In the unconstrained cases, DFNO shows dominant performance over all baseline models in all evaluation metrics for 2 times downscaling on which it was trained. This performance advantage persists when the DFNO model trained on 2 times downscaling is asked to zero-shot generalize to 4 times downscaling, where it outperforms models directly trained on the 4 times downscaling dataset such as CNN-4 and GAN-4. After the constraint layer is applied, DFNO's skill is further boosted for both 2 times and 4 times downscaling. It is consistent with the conclusion by \citet{Harder_2022} that training networks with the constraint layer can introduce an inductive bias, helping networks give more accurate downscaling results. However, note that in the zero-shot downscaling cases where bicubic interpolation is used to adjust network output resolution (i.e., 4 times downscaling with CNN-2, 2 times downscaling with CNN-4, 4 times downscaling with GAN-2, and 2 times downscaling with GAN-4), the constraint layer generally degrades model performance. This is probably because these networks are not trained to adapt to the renormalization operation inside the constraint layer with a different upsampling factor.  

%Figure \ref{fig:PDE_upsample_metric} visualizes results from Tables \ref{tab:PDE_upsample_unconstrained} and \ref{tab:PDE_upsample_constrained}. It confirms the superiority of DFNO in either learned or zero-shot downscaling; and verifies the effectiveness of constraint layers for improving model downscaling skills. 

\begin{figure}[ht]
    \centering
    \includegraphics[width=\textwidth]{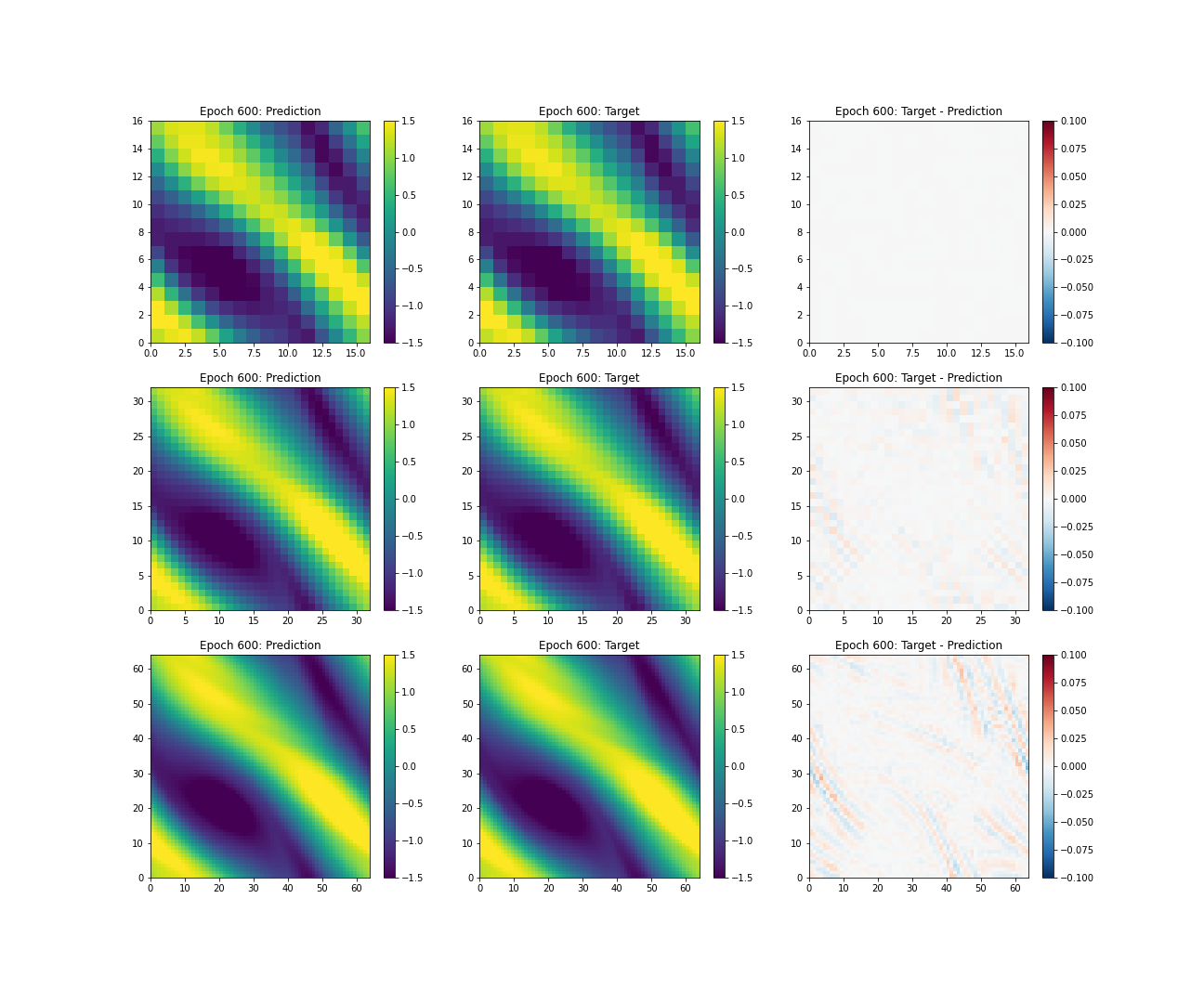}
    \caption{This figure shows the downscaling performance of our DFNO model with softmax constraint layer on the PDE solution data. The DFNO model was trained with 2 times downscaling data, then evaluated at 1 time (row 1), 2 times (row 2), and 4 times (row 3) downscaling. Column 1 shows the outputs from our DFNO model; column 2 is the numerical solution ground truth; and the difference (one order of magnitude lower than the ground truth values) between truth and prediction is presented in column 3.}
    \label{fig:PDE_upsample_constrained}
\end{figure}

One PDE solution downscaling example by our constrained DFNO model is presented in Figure \ref{fig:PDE_upsample_constrained}. The top row shows the result of input reconstruction (1 time downscaling). Because of the softmax constraint layer, DFNO trivially reconstructs the exact input because the conservation law enforces the output to equal the input when the upsampling factor is 1. Rows 2 and 3 illustrate 2 times (learned) and 4 times (zero-shot) downscaling results by DFNO. In both cases, the downscaled outputs (column 1) are very close to the ground truth (column 2), and the difference (column 3) is minor and negligible with values roughly one order of magnitude lower than the ground truth values.

\subsection{Downscaling ERA5 Climate Data}
The ERA5 climate and weather dataset \citep{Hersbach_2020} is a reanalysis product from the European Center for Medium-Range Weather Forecasts (ECMWF) that combines model data with worldwide observations. The observations are used as boundary conditions for numerical models that then predict various atmospheric variables. ERA5 is available as global hourly data with a $0.25^{\circ}\times0.25^{\circ}$ resolution, which is roughly $25~\mbox{km}$ per pixel. It covers all years starting from 1950. 

For this work, the quantity we focus on is the total column water that describes the vertical integral of the total amount of atmospheric water content, including water vapor, cloud water, and cloud ice but not precipitation. At each time step, we extract a random $128\times128$ patch from the global water content field of size $721\times1440$. There are roughly 60,000 time steps available in total. From these, 40,000 patches are randomly sampled for training and 10,000 for each validation and testing. The low-resolution counterparts are created by average pooling on high-resolution samples following the standard practice as in \citet{Serifi_2021, Leinonen_2021}. It results in low-resolution samples of sizes $32 \times 32$ and $64 \times 64$. This operation is physically sound, considering that conservation of water content means that the water content (density per squared meter) described in a low-resolution pixel should be equal to the average of its corresponding high-resolution pixels.

\begin{figure}[htb]
    \centering
    \includegraphics[width=\textwidth]{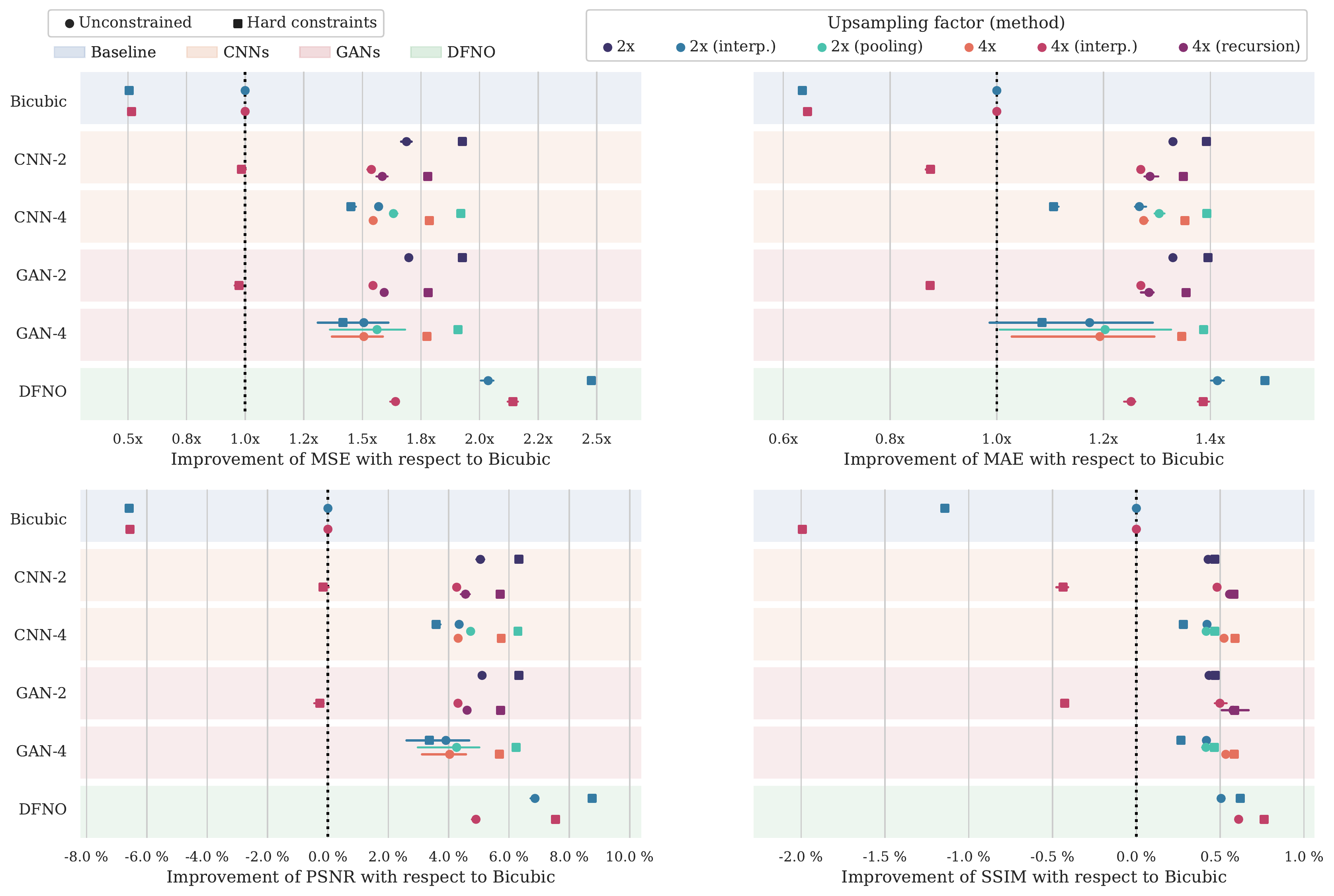}
    \caption{Metrics for downscaling models applied to the ERA5 dataset. Downscaling models CNN-2 (CNN-4) and GAN-2 (GAN-4) are trained with 2 times (4 times) downscaling data; the DFNO model is only trained with 2 times downscaling data. Each downscaling model is evaluated on both 2 times and 4 times downscaling. The 2 times downscaling outputs by CNN-2 and GAN-2 increase their resolution to 4 times through model recursion and bicubic interpolation. The 4 times downscaling outputs by CNN-4 and GAN-4 decrease their resolution to 2 times through average pooling and bicubic interpolation. Square (dot) denotes constrained (unconstrained) models. The metric mean and confidence interval from 3 runs are shown relatively to unconstrained bicubic interpolation. Model performance is evaluated by comparing marks of the same upsampling factor: cold colors for 2 times and warm colors for 4 times. Metric numerics and more details can be found in Tables \ref{tab:ERA5_upsample_unconstrained_interp}, \ref{tab:ERA5_upsample_unconstrained_other}, \ref{tab:ERA5_upsample_constrained_interp}, and \ref{tab:ERA5_upsample_constrained_other}.}
    \label{fig:ERA5_upsample_metric}
\end{figure}

As in the previous section, a DFNO model is trained with 2 times downscaling data and tested at 1 times, 2 times, and 4 times downscaling. Its performance is also compared against two CNN and two GAN downscaling models of upsampling factors 2 and 4. To enforce conservation law, a separate set of experiments are conducted with the softmax constraint layer applied. The downscaling performance of all models is collected in Tables \ref{tab:ERA5_upsample_unconstrained_interp} and \ref{tab:ERA5_upsample_unconstrained_other} (without constraint layer) and Tables \ref{tab:ERA5_upsample_constrained_interp} and \ref{tab:ERA5_upsample_constrained_other} (with constraint layer), and we provide a visualization of the relative improvement with respect to the bicubic baseline in Figure \ref{fig:ERA5_upsample_metric}. 

When the constraint layer is not applied, in learned (2 times) downscaling, we find that DFNO has the highest skill among all baseline models in all evaluation metrics. For zero-shot (4 times) downscaling, DFNO has an MAE score slightly worse than baseline CNN-2, CNN-4 and GAN-2 models but shows performance dominance in all the other metrics. Better performance in terms of MSE than MAE means the DFNO prediction errors mostly concentrate at values with magnitude less than 1. It is likely due to the fact that our DFNO is trained with MSE as the loss function, which is more sensitive to errors with large magnitude. After applying the constraint layer, DFNO performance in both learned and zero-shot downscaling is boosted showing performance dominance for all metrics. 
%Similarly to the previous section, for non-DFNO deep learning models, the constraint layer improves performance only when the model is trained with it. It results from the fact that those models by design are unable to adapt to unseen upsampling factor. 

%capture the high-frequency information presented in observational data. On the other hand, climate simulation data based on PDEs tend to be smoother and have little energy at high-frequency components.

%We attribute this to the fact that PDE solutions are much smoother than water content fields meaning there is more energy at high-frequency components in water content data due to measurement noise. On the other hand, FNO applies transformation on a truncated Fourier series, which makes it impossible for FNO to capture those high-frequency features demonstrated in water content data.

\begin{figure}[ht]
    \centering
    \includegraphics[width=\textwidth]{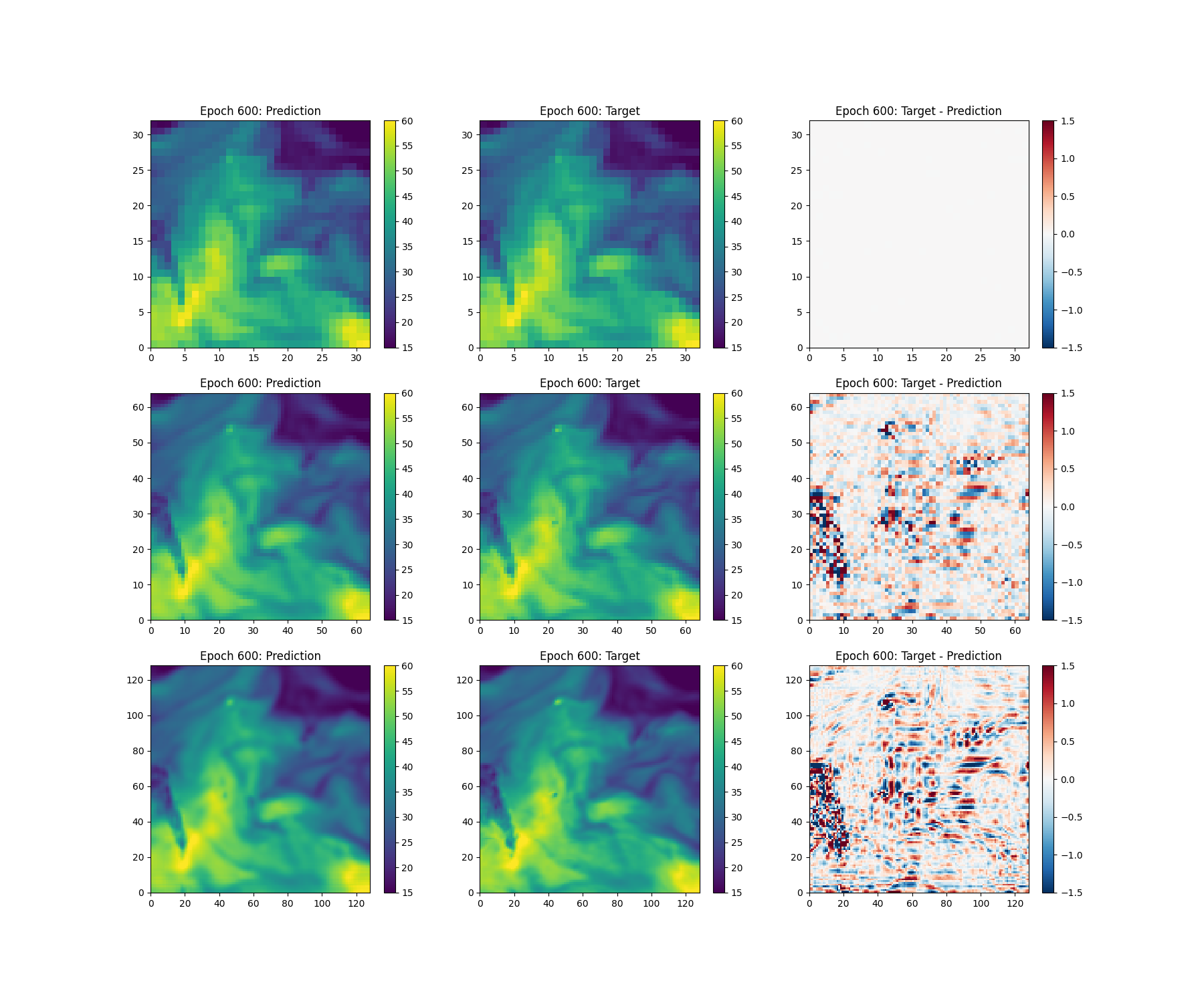}
    \caption{This figure shows the downscaling performance of our DFNO model with softmax constraint layer on ERA5 water content data. The DFNO model was trained with 2 times downscaling data, then evaluated at 1 time (row 1), 2 times (row 2), and 4 times (row 3) downscaling. Column 1 shows the outputs from our DFNO model; column 2 is the ground truth; the difference between truth and prediction is presented in column 3.}
    \label{fig:ERA5_upsample_constrained}
\end{figure}

Figure \ref{fig:ERA5_upsample_constrained} illustrates a case study on constrained DFNO downscaling ERA5 water content data. The softmax constraint layer helps DFNO reconstruct input perfectly (row 1). The 2 times downscaled (row 2) and 4 times downscaled (row 3) outputs are visually close to the ground truth (column 2) and with rather high perceptual quality as validated by quantitative metric scores in Tables \ref{tab:ERA5_upsample_constrained_interp} and \ref{tab:ERA5_upsample_constrained_other}. However, the error in column 3 is not as small as in the case of PDE solution downscaling (Figure \ref{fig:PDE_upsample_constrained}). It is not surprising as our model is intended for PDE based climate simulation data downscaling rather than observational climate data downscaling; the FNO inside our model applies transformations on a truncated Fourier series, so it is naturally easier for it to model simulation data which have a more succinct representation in Fourier basis than observational data.  

\subsection{Downscaling for PDE Integration}
This section considers the use of DFNO in integrating PDEs at high resolution (i.e., generating high resolution PDE solutions). There has been increasing interest in the use of data-driven deep learning-based methods to predict PDE solutions autoregressively \citep{li2021Fourier}, and the Fourier neural operator was introduced as a state-of-the-art approach in this regard. Here, we show that the DFNO paradigm has the potential to significantly improve upon the standard FNO approach. Namely, we assume that we have access to an accurate \emph{low-resolution} PDE solver, then use the DFNO to downscale the solution to higher resolution. Having a low-resolution PDE solution is a plausible assumption, since traditional numerical solvers are prohibitively time-intensive at high resolution but can be very cheap to run at low resolution.

Here we consider the Navier-Stokes equation as in Section \ref{sec:PDE_downscaling}. In the previous two sections, we have seen that the constrained DFNO outperforms the unconstrained DFNO; as a result, we here use only the constrained model. We train two different DFNO models, using 2 times ($16 \times 16 \rightarrow 32 \times 32$) and 4 times ($16 \times 16 \rightarrow 64 \times 64$) PDE downscaling data, respectively, and denoted as DFNO-2 and DFNO-4. We compare our approach against the standard FNO method, which predicts a solution one time step forward based on the solution at the previous ten time steps; two FNO models are trained with solution data at resolution $32 \times 32$ and $64 \times 64$, and are denoted as FFNO-32 and FFNO-64. Becasue FFNO-32 and FFNO-64 are resolution invariant, both of them are tested solving the Navier-Stokes equation at resolutions $32 \times 32$ and $64 \times 64$. In the end, all four models are evaluated with generated PDE solutions at resolutions $32 \times 32$ and $64 \times 64$.

\begin{figure}[htb]
    \centering
    \includegraphics[width=0.85\textwidth]{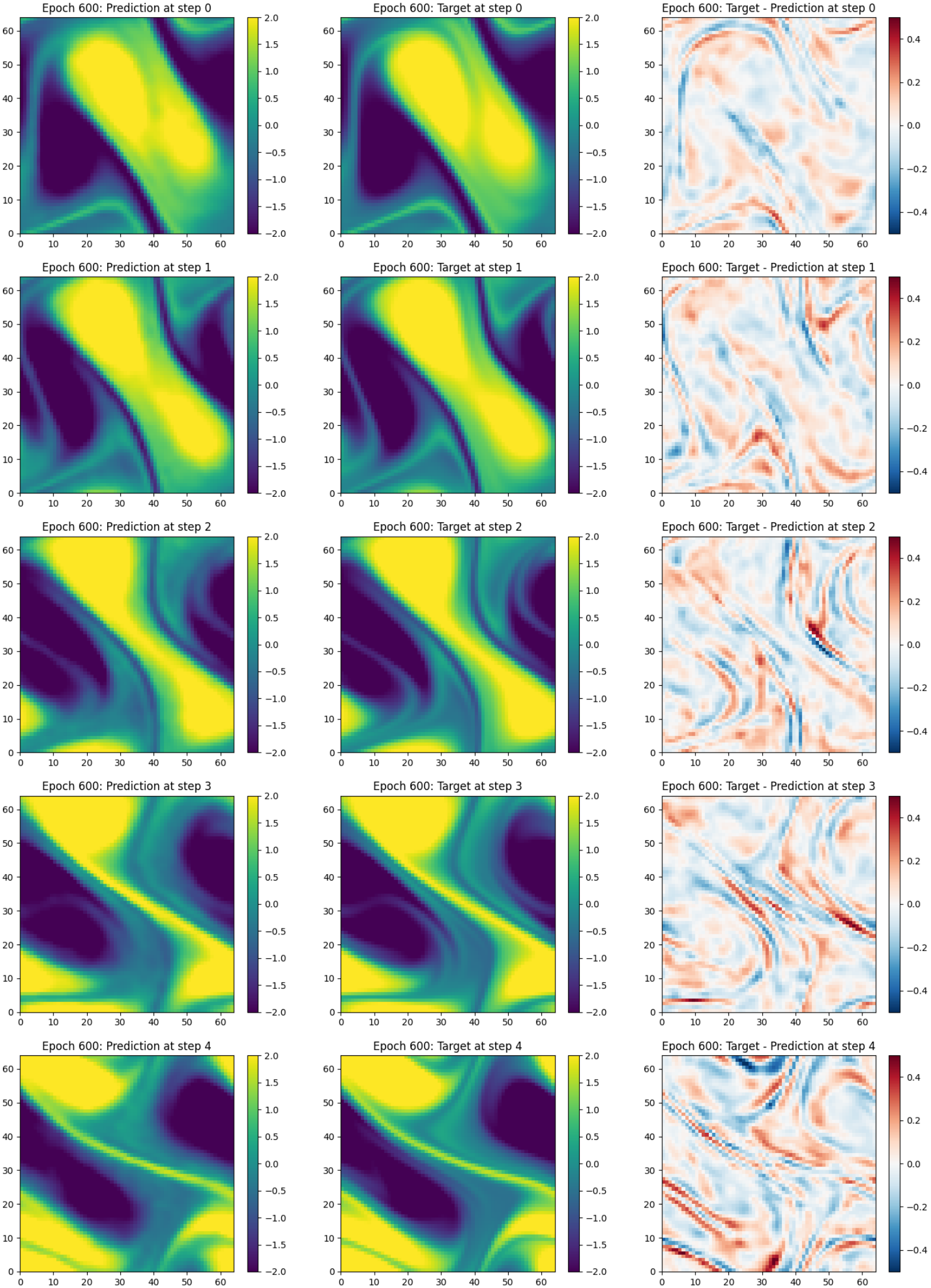}
    \caption{This figure shows Navier-Stokes equation solution $(64 \times 64)$ at five consecutive time steps (row 1 to row 5). The solution is generated by FFNO-64, a forward solution prediction model trained on a solution dataset of resolution $64 \times 64$. Column 1 shows FFNO-64 predicted solution; column 2 is the numerical solution ground truth; column 3 shows the difference between column 1 and column 2.}
    \label{fig:PDE_FFNO_64}
\end{figure}

\begin{figure}[htb]
    \centering
    \includegraphics[width=0.85\textwidth]{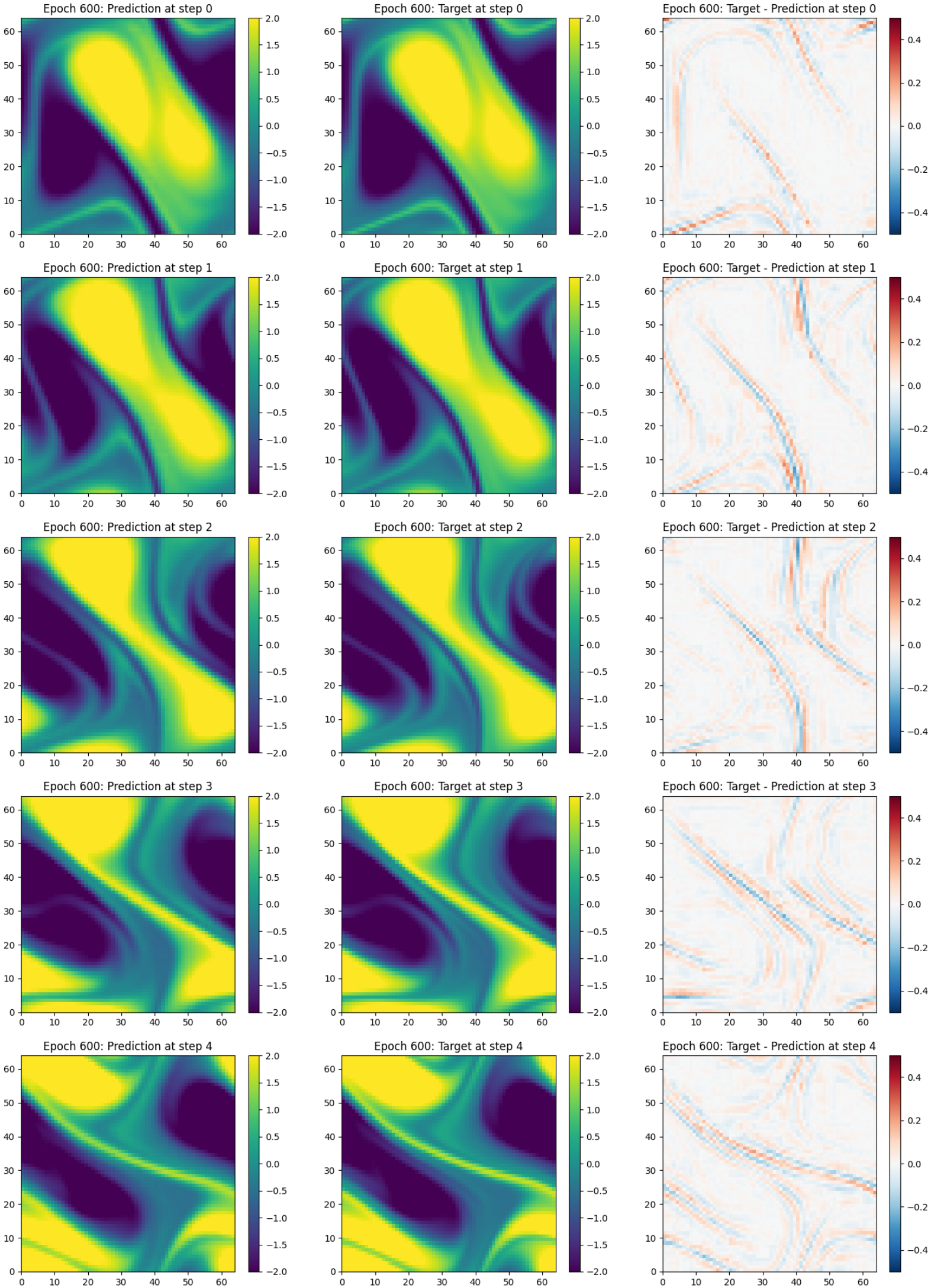}
    \caption{Similar to Figure \ref{fig:PDE_FFNO_64} but the solution is generated by DFNO-2. It is a constrained DFNO model trained on solution downscaling data from $16 \times 16$ to $32 \times 32$. It performs zero-shot downscaling on a solution from $16 \times 16$ to $64 \times 64$.}
    \label{fig:PDE_2x_DFNO_64}
\end{figure}

The solutions generated by DFNO and FFNO models are compared against ground truth numerical solutions, and the performance is summarized in Table \ref{tab:PDE_integration}. Overall, DFNO models show a significant performance advantage over FFNO models. Comparing between DFNO models, it is not surprising that zero-shot downscaling is still not as good as learned downscaling. To evaluate DFNO and FFNO performance visually, solution examples generated by FFNO-64 and DFNO-2 at resolution $64 \times 64$ for five consecutive time steps are presented in Figures \ref{fig:PDE_FFNO_64} and \ref{fig:PDE_2x_DFNO_64}. The generated solutions (column 1) are very close to numerical solutions (column 2) for both models. On the other hand, even though DFNO-2 zero-shot results are compared against FFNO-64 learned results, the error magnitude by DFNO-2 (column 3) is much less than that by FFNO-64. Results from both quantitative metric scores and visual illustration demonstrate that downscaling low-resolution solutions from numerical solvers gives better accuracy than generating by data-driven high-resolution solvers, that is, inputting low-resolution solutions as guidance makes it much easier to generate realistic high-resolution solutions.
%DFNO can zero-shot downscale low-resolution solutions to an unseen high resolution with reasonable quality.

\begin{table}[htb] 
\caption{This table compares two ways of solving the Navier-Stokes equation at high resolution concerning mean squared error (MSE) and mean absolute error (MAE). First: solve the equation numerically at low resolution ($16 \times 16$), then downscale the solution to $32 \times 32$ and $64 \times 64$ by constrained DFNO models. Second: use data-driven FFNO models to auto-regressively predict solutions at resolution $32 \times 32$ and $64 \times 64$.}
\label{tab:PDE_integration}
\begin{center}
\begin{tabular}{cccccc}
\toprule
Metric & Resolution & DFNO-2 & DFNO-4 & FFNO-32 & FFNO-64\\
\midrule
MSE & $32 \times 32$ & \bf{\color{red}0.0004} & \bf{\color{blue}0.0012} & 0.0101 & 0.0113 \\
MSE & $64 \times 64$ & \bf{\color{blue}0.0018} & \bf{\color{red}0.0007} & 0.0136 & 0.0118 \\
\midrule
MAE & $32 \times 32$ & \bf{\color{red}0.0124} & \bf{\color{blue}0.0208} & 0.0677 & 0.0725 \\
MAE & $64 \times 64$ & \bf{\color{blue}0.0246} & \bf{\color{red}0.0168} & 0.0788 & 0.0739 \\
\bottomrule
\end{tabular}
\end{center}
\end{table}

\section{Conclusion}
%In this study, we design the first arbitrary resolution downscaling model for climate data using Fourier neural operator.
In this work, we introduce the first arbitrary resolution downscaling model for climate data. This model takes in a low-resolution sample and outputs a function that interpolates the observed high-resolution counterpart. The low-resolution input is downscaled to an arbitrarily high resolution by evaluating the output function at discrete points. This model consists of three components: neural network, discretization inversion operator, and neural operator. They are implemented respectively as a residual convolutional network, bicubic interpolation, and a Fourier neural operator.

Our model is evaluated on a Navier-Stokes equation solution dataset and an ERA5 reanalysis water content dataset. It improves downscaling performance on both datasets significantly relative to state-of-the-art CNN and GAN super-resolution methods. It also zero-shot generalizes to higher upsampling factors, outperforming models directly trained on those factors. Our model's performance is further boosted when a softmax constraint layer is applied to enforce conservation laws. Finally, we compare two ways to integrate PDEs at high resolution. Combining our downscaling model with a low-resolution numerical solver, the downscaled solution has superior accuracy to that of the state-of-the-art high-resolution data-driven solver. 

While our DFNO approach conveys a significant performance improvement across all tasks, it demonstrates an even greater efficacy on climate simulation (Navier-Stokes equation) data as compared to observational climate (total water content) data. This may result from the fact that simulation data are much smoother than observational data; that is, simulation data have a more succinct representation in the Fourier basis than observational data. Therefore, simulation data are easier to be captured by a Fourier neural operator with a truncated Fourier series. It would be interesting to explore how to modify our model to adapt to data without a succinct Fourier representation so that its performance on observational climate data can be further improved.

% Acknowledgements and Disclosure of Funding should go at the end, before appendices and references

\acks{This work was supported in part by the Québec Ministère de l'Économie et de l'Innovation, IBM, and the Canada CIFAR AI Chairs Program. The authors also acknowledge material support from NVIDIA in the form of computational resources, and are grateful for technical support from the Mila IDT team in maintaining the Mila Compute Cluster.}

% Manual newpage inserted to improve layout of sample file - not
% needed in general before appendices/bibliography.

\newpage
\section*{Appendix}
\begin{table}[htb] 
\caption{Downscaling performance on the PDE dataset in terms of mean squared error (MSE), mean absolute error (MAE), peak signal-to-noise ratio (PSNR), and structural similarity index measure (SSIM). The best scores are highlighted in bold red, second best in bold blue. The DFNO model was trained on 2 times downscaling data, then tested on 1 time, 2 times, and 4 times downscaling. CNN-2 (GAN-2) and CNN-4 (GAN-2) represent convolutional (generative adversarial) downscaling models with predefined upsampling factors 2 and 4. They were trained on datasets of their corresponding upsampling factors, whose downscaling results are then downsampled or upsampled via bicubic interpolation to get desired resolution for evaluation.}
\label{tab:PDE_upsample_unconstrained_interp}
\begin{center}
\begin{tabular}{cccccccc}
\toprule
Metric & Factor & DFNO & CNN-2 & CNN-4 & GAN-2 & GAN-4 & Bicubic \\
\midrule
MSE & $1 \times$ & 0.0146 & 0.0057 & 0.0123 & \bf{\color{blue}0.0056} & 0.0131 & \bf{\color{red}0.0000} \\
MSE & $2 \times$ & \bf{\color{red}0.0015} & \bf{\color{blue}0.0042} & 0.0052 & 0.0044 & 0.0062 & 0.0252 \\
MSE & $4 \times$ & \bf{\color{red}0.0037} & 0.0093 & \bf{\color{blue}0.0070} & 0.0095 & 0.0080 & 0.0350 \\
\midrule
MAE & $1 \times$ & 0.0826 & 0.0524 & 0.0697 & \bf{\color{blue}0.0520} & 0.0746 & \bf{\color{red}0.0000} \\
MAE & $2 \times$ & \bf{\color{red}0.0238} & \bf{\color{blue}0.0397} & 0.0458 & 0.0424 & 0.0534 & 0.1027 \\
MAE & $4 \times$ & \bf{\color{red}0.0359} & 0.0579 & \bf{\color{blue}0.0495} & 0.0601 & 0.0573 & 0.1150 \\
\midrule
PSNR & $1 \times$ & 40.2750 & 44.3504 & 41.0302 & \bf{\color{blue}44.4541} & 40.7810 & \bf{\color{red}154.0983} \\
PSNR & $2 \times$ & \bf{\color{red}50.2061} & \bf{\color{blue}45.7778} & 44.8762 & 45.5806 & 44.2337 & 38.0326 \\
PSNR & $4 \times$ & \bf{\color{red}46.3361} & 42.4054 & \bf{\color{blue}43.6083} & 42.3192 & 43.1123 & 36.6248 \\
\midrule
SSIM & $1 \times$ & 0.9934 & \bf{\color{blue}0.9968} & 0.9935 & 0.9963 & 0.9890 & \bf{\color{red}1.0000} \\
SSIM & $2 \times$ & \bf{\color{red}0.9981} & \bf{\color{blue}0.9963} & 0.9952 & 0.9956 & 0.9917 & 0.9741 \\
SSIM & $4 \times$ & \bf{\color{red}0.9920} & 0.9842 & \bf{\color{blue}0.9879} & 0.9835 & 0.9847 & 0.9335 \\
\bottomrule
\end{tabular}
\end{center}
\end{table}

\newpage
\begin{table}[htb] 
\caption{Similar to Table \ref{tab:PDE_upsample_unconstrained_interp} but softmax constraint layer is applied to the output of each model.}
\label{tab:PDE_upsample_constrained_interp}
\begin{center}
\begin{tabular}{cccccccc}
\toprule
Metric & Factor & DFNO & CNN-2 & CNN-4 & GAN-2 & GAN-4 & Bicubic \\
\midrule
MSE & $1 \times$ & \bf{\color{red}0.0000} & \bf{\color{red}0.0000} & \bf{\color{red}0.0000} & \bf{\color{red}0.0000} & \bf{\color{red}0.0000} & \bf{\color{red}0.0000} \\
MSE & $2 \times$ & \bf{\color{red}0.0011} & \bf{\color{blue}0.0038} & 0.0063 & 0.0038 & 0.0084 & 0.0365 \\
MSE & $4 \times$ & \bf{\color{red}0.0029} & 0.0217 & \bf{\color{blue}0.0063} & 0.0228 & 0.0064 & 0.0517 \\
\midrule
MAE & $1 \times$ & \bf{\color{red}0.0000} & \bf{\color{red}0.0000} & \bf{\color{red}0.0000} & \bf{\color{red}0.0000} & \bf{\color{red}0.0000} & \bf{\color{red}0.0000} \\
MAE & $2 \times$ & \bf{\color{red}0.0196} & \bf{\color{blue}0.0363} & 0.0528 & 0.0365 & 0.0627 & 0.1241 \\
MAE & $4 \times$ & \bf{\color{red}0.0313} & 0.1032 & \bf{\color{blue}0.0457} & 0.1058 & 0.0462 & 0.1431 \\
\midrule
PSNR & $1 \times$ & 151.8861 & \bf{\color{red}153.3908} & 152.4238 & \bf{\color{blue}153.3476} & 152.1304 & 152.4239 \\
PSNR & $2 \times$ & \bf{\color{red}51.8071} & \bf{\color{blue}46.2719} & 44.2463 & 46.2266 & 43.0041 & 36.4336 \\
PSNR & $4 \times$ & \bf{\color{red}47.4375} & 38.7146 & \bf{\color{blue}44.1036} & 38.5096 & 44.0425 & 34.9377 \\
\midrule
SSIM & $1 \times$ & \bf{\color{red}1.0000} & \bf{\color{red}1.0000} & \bf{\color{red}1.0000} & \bf{\color{red}1.0000} & \bf{\color{red}1.0000} & \bf{\color{red}1.0000} \\
SSIM & $2 \times$ & \bf{\color{red}0.9987} & \bf{\color{blue}0.9969} & 0.9942 & 0.9969 & 0.9920 & 0.9659 \\
SSIM & $4 \times$ & \bf{\color{red}0.9937} & 0.9605 & \bf{\color{blue}0.9894} & 0.9583 & 0.9892 & 0.9108 \\
\bottomrule
\end{tabular}
\end{center}
\end{table}

\newpage
\begin{table}[htb] 
\caption{Downscaling performance on the PDE dataset in terms of mean squared error (MSE), mean absolute error (MAE), peak signal-to-noise ratio (PSNR), and structural similarity index measure (SSIM). The best scores are highlighted in bold red, second best in bold blue. The DFNO model was trained on 2 times downscaling data, then tested on 1 time, 2 times, and 4 times downscaling. CNN-2 (GAN-2) and CNN-4 (GAN-2) represent convolutional (generative adversarial) downscaling models with predefined upsampling factors 2 and 4. They were trained on datasets of their corresponding upsampling factors, whose downscaling results are then downsampled (upsampled) via average pooling (model recursion) to get desired resolution for evaluation.}
\label{tab:PDE_upsample_unconstrained_other}
\begin{center}
\begin{tabular}{cccccccc}
\toprule
Metric & Factor & DFNO & CNN-2 & CNN-4 & GAN-2 & GAN-4 & Bicubic \\
\midrule
MSE & $1 \times$ & 0.0146 & 0.0002 & \bf{\color{blue}0.0002} & 0.0004 & 0.0011 & \bf{\color{red}0.0000} \\
MSE & $2 \times$ & \bf{\color{red}0.0015} & 0.0042 & \bf{\color{blue}0.0042} & 0.0044 & 0.0051 & 0.0252 \\
MSE & $4 \times$ & \bf{\color{red}0.0037} & 0.0076 & \bf{\color{blue}0.0070} & 0.0083 & 0.0080 & 0.0350 \\
\midrule
MAE & $1 \times$ & 0.0826 & \bf{\color{blue}0.0097} & 0.0098 & 0.0164 & 0.0237 & \bf{\color{red}0.0000} \\
MAE & $2 \times$ & \bf{\color{red}0.0238} & 0.0397 & \bf{\color{blue}0.0394} & 0.0424 & 0.0477 & 0.1027 \\
MAE & $4 \times$ & \bf{\color{red}0.0359} & 0.0535 & \bf{\color{blue}0.0495} & 0.0600 & 0.0573 & 0.1150 \\
\midrule
PSNR & $1 \times$ & 40.2750 & \bf{\color{blue}60.8445} & 60.4059 & 56.6575 & 55.8089 & \bf{\color{red}154.0983} \\
PSNR & $2 \times$ & \bf{\color{red}50.2061} & 45.7778 & \bf{\color{blue}45.8595} & 45.5806 & 45.1245 & 38.0326 \\
PSNR & $4 \times$ & \bf{\color{red}46.3361} & 43.2835 & \bf{\color{blue}43.6083} & 42.8902 & 43.1123 & 36.6248 \\
\midrule
SSIM & $1 \times$ & 0.9934 & \bf{\color{blue}0.9996} & 0.9996 & 0.9989 & 0.9953 & \bf{\color{red}1.0000} \\
SSIM & $2 \times$ & \bf{\color{red}0.9981} & 0.9963 & \bf{\color{blue}0.9963} & 0.9956 & 0.9928 & 0.9741 \\
SSIM & $4 \times$ & \bf{\color{red}0.9920} & 0.9868 & \bf{\color{blue}0.9879} & 0.9849 & 0.9847 & 0.9335 \\
\bottomrule
\end{tabular}
\end{center}
\end{table}

\newpage
\begin{table}[htb] 
\caption{Similar to Table \ref{tab:PDE_upsample_unconstrained_other} but softmax constraint layer is applied to the output of each model.}
\label{tab:PDE_upsample_constrained_other}
\begin{center}
\begin{tabular}{cccccccc}
\toprule
Metric & Factor & DFNO & CNN-2 & CNN-4 & GAN-2 & GAN-4 & Bicubic \\
\midrule
MSE & $1 \times$ & \bf{\color{red}0.0000} & \bf{\color{red}0.0000} & \bf{\color{red}0.0000} & \bf{\color{red}0.0000} & \bf{\color{red}0.0000} & \bf{\color{red}0.0000} \\
MSE & $2 \times$ & \bf{\color{red}0.0011} & 0.0038 & \bf{\color{blue}0.0036} & 0.0038 & 0.0036 & 0.0365 \\
MSE & $4 \times$ & \bf{\color{red}0.0029} & 0.0067 & \bf{\color{blue}0.0063} & 0.0067 & 0.0064 & 0.0517 \\
\midrule
MAE & $1 \times$ & \bf{\color{red}0.0000} & \bf{\color{red}0.0000} & \bf{\color{red}0.0000} & \bf{\color{red}0.0000} & \bf{\color{red}0.0000} & \bf{\color{red}0.0000} \\
MAE & $2 \times$ & \bf{\color{red}0.0196} & 0.0363 & \bf{\color{blue}0.0354} & 0.0365 & 0.0357 & 0.1241 \\
MAE & $4 \times$ & \bf{\color{red}0.0313} & 0.0474 & \bf{\color{blue}0.0457} & 0.0478 & 0.0462 & 0.1431 \\
\midrule
PSNR & $1 \times$ & \bf{\color{blue}151.8861} & 149.9829 & 147.8327 & 149.3479 & 147.5434 & \bf{\color{red}152.4239} \\
PSNR & $2 \times$ & \bf{\color{red}51.8071} & 46.2719 & \bf{\color{blue}46.5235} & 46.2266 & 46.4569 & 36.4336 \\
PSNR & $4 \times$ & \bf{\color{red}47.4375} & 43.8382 & \bf{\color{blue}44.1036} & 43.7889 & 44.0425 & 34.9377 \\
\midrule
SSIM & $1 \times$ & \bf{\color{red}1.0000} & \bf{\color{red}1.0000} & \bf{\color{red}1.0000} & \bf{\color{red}1.0000} & \bf{\color{red}1.0000} & \bf{\color{red}1.0000} \\
SSIM & $2 \times$ & \bf{\color{red}0.9987} & 0.9969 & \bf{\color{blue}0.9971} & 0.9969 & 0.9970 & 0.9659 \\
SSIM & $4 \times$ & \bf{\color{red}0.9937} & 0.9889 & \bf{\color{blue}0.9894} & 0.9887 & 0.9892 & 0.9108 \\
\bottomrule
\end{tabular}
\end{center}
\end{table}

\newpage
\begin{table}[htb] 
\caption{Downscaling performance on the ERA5 water content dataset in terms of mean squared error (MSE), mean absolute error (MAE), peak signal-to-noise ratio (PSNR), and structural similarity index measure (SSIM). The best scores are highlighted in bold red, second best in bold blue. The DFNO model was trained on 2 times downscaling data, then tested on 1 time, 2 times, and 4 times downscaling. CNN-2 (GAN-2) and CNN-4 (GAN-2) represent convolutional (generative adversarial) downscaling models with predefined upsampling factors 2 and 4. They were trained on datasets of their corresponding upsampling factors, whose downscaling results are then downsampled or upsampled via bicubic interpolation to get desired resolution for evaluation.}
\label{tab:ERA5_upsample_unconstrained_interp}
\begin{center}
\begin{tabular}{cccccccc}
\toprule
Metric & Factor & DFNO & CNN-2 & CNN-4 & GAN-2 & GAN-4 & Bicubic \\
\midrule
MSE & $1 \times$ & 0.2140 & 0.0940 & 0.1566 & \bf{\color{blue}0.0930} & 0.1752 & \bf{\color{red}0.0000} \\
MSE & $2 \times$ & \bf{\color{red}0.2063} & 0.2488 & 0.2677 & \bf{\color{blue}0.2474} & 0.2815 & 0.4201 \\
MSE & $4 \times$ & \bf{\color{red}0.3628} & 0.3870 & \bf{\color{blue}0.3851} & 0.3853 & 0.3970 & 0.5954 \\
\midrule
MAE & $1 \times$ & 0.2896 & 0.1737 & 0.2149 & \bf{\color{blue}0.1731} & 0.2439 & \bf{\color{red}0.0000} \\
MAE & $2 \times$ & \bf{\color{red}0.2392} & \bf{\color{blue}0.2541} & 0.2668 & 0.2541 & 0.2920 & 0.3380 \\
MAE & $4 \times$ & 0.3067 & 0.3023 & \bf{\color{red}0.3010} & \bf{\color{blue}0.3022} & 0.3251 & 0.3838 \\
\midrule
PSNR & $1 \times$ & 46.9630 & 50.5294 & 48.3152 & \bf{\color{blue}50.5795} & 47.8863 & \bf{\color{red}173.5160} \\
PSNR & $2 \times$ & \bf{\color{red}48.1002} & 47.2860 & 46.9688 & \bf{\color{blue}47.3110} & 46.7714 & 45.0115 \\
PSNR & $4 \times$ & \bf{\color{red}46.0154} & 45.7349 & \bf{\color{blue}45.7560} & 45.7535 & 45.6334 & 43.8633 \\
\midrule
SSIM & $1 \times$ & 0.9964 & 0.9982 & 0.9971 & \bf{\color{blue}0.9982} & 0.9971 & \bf{\color{red}1.0000} \\
SSIM & $2 \times$ & \bf{\color{red}0.9941} & 0.9933 & 0.9933 & \bf{\color{blue}0.9934} & 0.9932 & 0.9891 \\
SSIM & $4 \times$ & \bf{\color{red}0.9895} & 0.9882 & 0.9886 & 0.9884 & \bf{\color{blue}0.9887} & 0.9835 \\
\bottomrule
\end{tabular}
\end{center}
\end{table}

\newpage
\begin{table}[htb] 
\caption{Similar to Table \ref{tab:ERA5_upsample_unconstrained_interp} but softmax constraint layer is applied to the output of each model.}
\label{tab:ERA5_upsample_constrained_interp}
\begin{center}
\begin{tabular}{cccccccc}
\toprule
Metric & Factor & DFNO & CNN-2 & CNN-4 & GAN-2 & GAN-4 & Bicubic \\
\midrule
MSE & $1 \times$ & \bf{\color{red}0.0000} & \bf{\color{red}0.0000} & \bf{\color{red}0.0000} & \bf{\color{red}0.0000} & \bf{\color{red}0.0000} & \bf{\color{red}0.0000} \\
MSE & $2 \times$ & \bf{\color{red}0.1696} & 0.2181 & 0.2896 & \bf{\color{blue}0.2181} & 0.2964 & 0.8314 \\
MSE & $4 \times$ & \bf{\color{red}0.2779} & 0.6054 & \bf{\color{blue}0.3334} & 0.6118 & 0.3355 & 1.1552 \\
\midrule
MAE & $1 \times$ & \bf{\color{red}0.0000} & \bf{\color{red}0.0000} & \bf{\color{red}0.0000} & \bf{\color{red}0.0000} & \bf{\color{red}0.0000} & \bf{\color{red}0.0000} \\
MAE & $2 \times$ & \bf{\color{red}0.2250} & 0.2427 & 0.3055 & \bf{\color{blue}0.2422} & 0.3116 & 0.5318 \\
MAE & $4 \times$ & \bf{\color{red}0.2768} & 0.4383 & \bf{\color{blue}0.2837} & 0.4386 & 0.2851 & 0.5950 \\
\midrule
PSNR & $1 \times$ & 164.1793 & \bf{\color{red}170.2039} & 166.2301 & \bf{\color{blue}169.6977} & 165.4083 & 161.0459 \\
PSNR & $2 \times$ & \bf{\color{red}48.9508} & 47.8584 & 46.6269 & \bf{\color{blue}47.8584} & 46.5268 & 42.0471 \\
PSNR & $4 \times$ & \bf{\color{red}47.1723} & 43.7915 & \bf{\color{blue}46.3820} & 43.7464 & 46.3549 & 40.9850 \\
\midrule
SSIM & $1 \times$ & \bf{\color{red}1.0000} & \bf{\color{red}1.0000} & \bf{\color{red}1.0000} & \bf{\color{red}1.0000} & \bf{\color{red}1.0000} & \bf{\color{red}1.0000} \\
SSIM & $2 \times$ & \bf{\color{red}0.9952} & 0.9937 & 0.9919 & \bf{\color{blue}0.9938} & 0.9917 & 0.9778 \\
SSIM & $4 \times$ & \bf{\color{red}0.9910} & 0.9792 & \bf{\color{blue}0.9893} & 0.9793 & 0.9892 & 0.9639 \\
\bottomrule
\end{tabular}
\end{center}
\end{table}

\newpage
\begin{table}[htb] 
\caption{Downscaling performance on the ERA5 water content dataset in terms of mean squared error (MSE), mean absolute error (MAE), peak signal-to-noise ratio (PSNR), and structural similarity index measure (SSIM). The best scores are highlighted in bold red, second best in bold blue. The DFNO model was trained on 2 times downscaling data, then tested on 1 time, 2 times, and 4 times downscaling. CNN-2 (GAN-2) and CNN-4 (GAN-2) represent convolutional (generative adversarial) downscaling models with predefined upsampling factors 2 and 4. They were trained on datasets of their corresponding upsampling factors, whose downscaling results are then downsampled (upsampled) via average pooling (model recursion) to get desired resolution for evaluation.}
\label{tab:ERA5_upsample_unconstrained_other}
\begin{center}
\begin{tabular}{cccccccc}
\toprule
Metric & Factor & DFNO & CNN-2 & CNN-4 & GAN-2 & GAN-4 & Bicubic \\
\midrule
MSE & $1 \times$ & 0.2140 & 0.0030 & 0.0046 & \bf{\color{blue}0.0028} & 0.0243 & \bf{\color{red}0.0000} \\
MSE & $2 \times$ & \bf{\color{red}0.2063} & 0.2488 & 0.2573 & \bf{\color{blue}0.2474} & 0.2713 & 0.4201 \\
MSE & $4 \times$ & \bf{\color{red}0.3628} & 0.3757 & 0.3851 & \bf{\color{blue}0.3737} & 0.3970 & 0.5954 \\
\midrule
MAE & $1 \times$ & 0.2896 & 0.0345 & 0.0440 & \bf{\color{blue}0.0339} & 0.1087 & \bf{\color{red}0.0000} \\
MAE & $2 \times$ & \bf{\color{red}0.2392} & \bf{\color{blue}0.2541} & 0.2592 & 0.2541 & 0.2852 & 0.3380 \\
MAE & $4 \times$ & 0.3067 & \bf{\color{red}0.2982} & 0.3010 & \bf{\color{blue}0.2987} & 0.3251 & 0.3838 \\
\midrule
PSNR & $1 \times$ & 46.9630 & 65.5137 & 63.8092 & \bf{\color{blue}65.8671} & 59.5288 & \bf{\color{red}173.5160} \\
PSNR & $2 \times$ & \bf{\color{red}48.1002} & 47.2860 & 47.1402 & \bf{\color{blue}47.3110} & 46.9310 & 45.0115 \\
PSNR & $4 \times$ & \bf{\color{red}46.0154} & 45.8635 & 45.7560 & \bf{\color{blue}45.8862} & 45.6334 & 43.8633 \\
\midrule
SSIM & $1 \times$ & 0.9964 & 1.0000 & 0.9999 & \bf{\color{blue}1.0000} & 0.9998 & \bf{\color{red}1.0000} \\
SSIM & $2 \times$ & \bf{\color{red}0.9941} & 0.9933 & 0.9932 & \bf{\color{blue}0.9934} & 0.9932 & 0.9891 \\
SSIM & $4 \times$ & \bf{\color{red}0.9895} & 0.9890 & 0.9886 & \bf{\color{blue}0.9892} & 0.9887 & 0.9835 \\
\bottomrule
\end{tabular}
\end{center}
\end{table}

\newpage
\begin{table}[htb] 
\caption{Similar to Table \ref{tab:ERA5_upsample_unconstrained_other} but softmax constraint layer is applied to the output of each model.}
\label{tab:ERA5_upsample_constrained_other}
\begin{center}
\begin{tabular}{cccccccc}
\toprule
Metric & Factor & DFNO & CNN-2 & CNN-4 & GAN-2 & GAN-4 & Bicubic \\
\midrule
MSE & $1 \times$ & \bf{\color{red}0.0000} & \bf{\color{red}0.0000} & \bf{\color{red}0.0000} & \bf{\color{red}0.0000} & \bf{\color{red}0.0000} & \bf{\color{red}0.0000} \\
MSE & $2 \times$ & \bf{\color{red}0.1696} & 0.2181 & 0.2188 & \bf{\color{blue}0.2181} & 0.2202 & 0.8314 \\
MSE & $4 \times$ & \bf{\color{red}0.2779} & 0.3347 & \bf{\color{blue}0.3334} & 0.3343 & 0.3355 & 1.1552 \\
\midrule
MAE & $1 \times$ & \bf{\color{red}0.0000} & \bf{\color{red}0.0000} & \bf{\color{red}0.0000} & \bf{\color{red}0.0000} & \bf{\color{red}0.0000} & \bf{\color{red}0.0000} \\
MAE & $2 \times$ & \bf{\color{red}0.2250} & 0.2427 & 0.2426 & \bf{\color{blue}0.2422} & 0.2436 & 0.5318 \\
MAE & $4 \times$ & \bf{\color{red}0.2768} & 0.2844 & 0.2837 & \bf{\color{blue}0.2833} & 0.2851 & 0.5950 \\
\midrule
PSNR & $1 \times$ & \bf{\color{red}164.1793} & 155.5239 & 153.5080 & 155.0793 & 153.3323 & \bf{\color{blue}161.0459} \\
PSNR & $2 \times$ & \bf{\color{red}48.9508} & 47.8584 & 47.8454 & \bf{\color{blue}47.8584} & 47.8165 & 42.0471 \\
PSNR & $4 \times$ & \bf{\color{red}47.1723} & 46.3649 & \bf{\color{blue}46.3820} & 46.3708 & 46.3549 & 40.9850 \\
\midrule
SSIM & $1 \times$ & \bf{\color{red}1.0000} & \bf{\color{red}1.0000} & \bf{\color{red}1.0000} & \bf{\color{red}1.0000} & \bf{\color{red}1.0000} & \bf{\color{red}1.0000} \\
SSIM & $2 \times$ & \bf{\color{red}0.9952} & 0.9937 & 0.9937 & \bf{\color{blue}0.9938} & 0.9937 & 0.9778 \\
SSIM & $4 \times$ & \bf{\color{red}0.9910} & 0.9892 & \bf{\color{blue}0.9893} & 0.9893 & 0.9892 & 0.9639 \\
\bottomrule
\end{tabular}
\end{center}
\end{table}

\newpage

\bibliography{reference}

\begin{thebibliography}{29}
\providecommand{\natexlab}[1]{#1}
\providecommand{\url}[1]{\texttt{#1}}
\expandafter\ifx\csname urlstyle\endcsname\relax
  \providecommand{\doi}[1]{doi: #1}\else
  \providecommand{\doi}{doi: \begingroup \urlstyle{rm}\Url}\fi

\bibitem[Balaji(2021)]{Balaji_2021}
V.~Balaji.
\newblock Climbing down charney's ladder: machine learning and the post-dennard
  era of computational climate science.
\newblock \emph{Philosophical Transactions of the Royal Society A:
  Mathematical, Physical and Engineering Sciences}, 379\penalty0
  (2194):\penalty0 20200085, 2021.
\newblock \doi{10.1098/rsta.2020.0085}.
\newblock URL
  \url{https://royalsocietypublishing.org/doi/abs/10.1098/rsta.2020.0085}.

\bibitem[Ballas et~al.(2015)Ballas, Yao, Pal, and Courville]{Ballas_2015}
N.~Ballas, L.~Yao, C.~Pal, and A.~Courville.
\newblock Delving deeper into convolutional networks for learning video
  representations, 2015.
\newblock URL \url{https://arxiv.org/abs/1511.06432}.

\bibitem[Beucler et~al.(2019)Beucler, Rasp, Pritchard, and
  Gentine]{Beucler_2019}
T.~Beucler, S.~Rasp, M.~Pritchard, and P.~Gentine.
\newblock Achieving conservation of energy in neural network emulators for
  climate modeling, 2019.
\newblock URL \url{https://arxiv.org/abs/1906.06622}.

\bibitem[Beucler et~al.(2021)Beucler, Pritchard, Rasp, Ott, Baldi, and
  Gentine]{Beucler_2021}
T.~Beucler, M.~Pritchard, S.~Rasp, J.~Ott, P.~Baldi, and P.~Gentine.
\newblock Enforcing analytic constraints in neural networks emulating physical
  systems.
\newblock \emph{Physical Review Letters}, 126\penalty0 (9), mar 2021.
\newblock \doi{10.1103/physrevlett.126.098302}.
\newblock URL \url{https://doi.org/10.1103%2Fphysrevlett.126.098302}.

\bibitem[Chaudhuri and Robertson(2020)]{Chaudhuri_2020}
C.~Chaudhuri and C.~Robertson.
\newblock {CliGAN}: A structurally sensitive convolutional neural network model
  for statistical downscaling of precipitation from multi-model ensembles.
\newblock \emph{Water}, 12\penalty0 (12):\penalty0 3353, nov 2020.
\newblock \doi{10.3390/w12123353}.
\newblock URL \url{https://doi.org/10.3390%2Fw12123353}.

\bibitem[Chen et~al.(2022)Chen, Feng, Liu, Ni, Lu, Tong, and Liu]{chen_2022}
X.~Chen, K.~Feng, N.~Liu, B.~Ni, Y.~Lu, Z.~Tong, and Z.~Liu.
\newblock Rainnet: A large-scale imagery dataset and benchmark for spatial
  precipitation downscaling, 2022.

\bibitem[Daw et~al.(2020)Daw, Thomas, Carey, Read, Appling, and
  Karpatne]{Daw_2020}
A.~Daw, R.~Q. Thomas, C.~C. Carey, J.~S. Read, A.~P. Appling, and A.~Karpatne.
\newblock Physics-guided architecture ({PGA}) of neural networks for
  quantifying uncertainty in lake temperature modeling.
\newblock In \emph{Proceedings of the 2020 {SIAM} International Conference on
  Data Mining}, pages 532--540. Society for Industrial and Applied Mathematics,
  jan 2020.
\newblock \doi{10.1137/1.9781611976236.60}.
\newblock URL \url{https://doi.org/10.1137%2F1.9781611976236.60}.

\bibitem[de~Boor(1962)]{de_Boor_1962}
C.~de~Boor.
\newblock Bicubic spline interpolation.
\newblock \emph{Journal of Mathematics and Physics}, 41\penalty0
  (1-4):\penalty0 212--218, apr 1962.
\newblock \doi{10.1002/sapm1962411212}.
\newblock URL \url{https://doi.org/10.1002%2Fsapm1962411212}.

\bibitem[Dong et~al.(2015)Dong, Loy, He, and Tang]{Dong_2015}
C.~Dong, C.~C. Loy, K.~He, and X.~Tang.
\newblock Image super-resolution using deep convolutional networks, 2015.
\newblock URL \url{https://arxiv.org/abs/1501.00092}.

\bibitem[Goodfellow et~al.(2014)Goodfellow, Pouget-Abadie, Mirza, Xu,
  Warde-Farley, Ozair, Courville, and Bengio]{Ian_2014}
I.~J. Goodfellow, J.~Pouget-Abadie, M.~Mirza, B.~Xu, D.~Warde-Farley, S.~Ozair,
  A.~Courville, and Y.~Bengio.
\newblock Generative adversarial networks, 2014.
\newblock URL \url{https://arxiv.org/abs/1406.2661}.

\bibitem[Groenke et~al.(2020)Groenke, Madaus, and Monteleoni]{Groenke_2020}
B.~Groenke, L.~Madaus, and C.~Monteleoni.
\newblock {ClimAlign}: Unsupervised statistical downscaling of climate
  variables via normalizing flows.
\newblock In \emph{Proceedings of the 10th International Conference on Climate
  Informatics}. {ACM}, sep 2020.
\newblock \doi{10.1145/3429309.3429318}.
\newblock URL \url{https://doi.org/10.1145%2F3429309.3429318}.

\bibitem[Harder et~al.(2022{\natexlab{a}})Harder, Watson-Parris, Stier,
  Strassel, Gauger, and Keuper]{Harder_2022_aerosol}
P.~Harder, D.~Watson-Parris, P.~Stier, D.~Strassel, N.~R. Gauger, and
  J.~Keuper.
\newblock Physics-informed learning of aerosol microphysics,
  2022{\natexlab{a}}.
\newblock URL \url{https://arxiv.org/abs/2207.11786}.

\bibitem[Harder et~al.(2022{\natexlab{b}})Harder, Yang, Ramesh, Sattigeri,
  Hernandez-Garcia, Watson, Szwarcman, and Rolnick]{Harder_2022}
P.~Harder, Q.~Yang, V.~Ramesh, P.~Sattigeri, A.~Hernandez-Garcia, C.~Watson,
  D.~Szwarcman, and D.~Rolnick.
\newblock Generating physically-consistent high-resolution climate data with
  hard-constrained neural networks, 2022{\natexlab{b}}.
\newblock URL \url{https://arxiv.org/abs/2208.05424}.

\bibitem[Harder et~al.(2023)Harder, Ramesh, Hernandez-Garcia, Yang, Sattigeri,
  Szwarcman, Watson, and Rolnick]{Harder_2023}
P.~Harder, V.~Ramesh, A.~Hernandez-Garcia, Q.~Yang, P.~Sattigeri, D.~Szwarcman,
  C.~Watson, and D.~Rolnick.
\newblock Physics-constrained deep learning for climate downscaling, 2023.

\bibitem[Hersbach et~al.(2020)Hersbach, Bell, Berrisford, Hirahara,
  Hor{\'a}nyi, Mu{\~n}oz-Sabater, Nicolas, Peubey, Radu, Schepers,
  et~al.]{Hersbach_2020}
H.~Hersbach, B.~Bell, P.~Berrisford, S.~Hirahara, A.~Hor{\'a}nyi,
  J.~Mu{\~n}oz-Sabater, J.~Nicolas, C.~Peubey, R.~Radu, D.~Schepers, et~al.
\newblock The era5 global reanalysis.
\newblock \emph{Quarterly Journal of the Royal Meteorological Society},
  146\penalty0 (730):\penalty0 1999--2049, 2020.

\bibitem[Höhlein et~al.(2020)Höhlein, Kern, Hewson, and
  Westermann]{H_hlein_2020}
K.~Höhlein, M.~Kern, T.~Hewson, and R.~Westermann.
\newblock A comparative study of convolutional neural network models for wind
  field downscaling.
\newblock \emph{Meteorological Applications}, 27\penalty0 (6), nov 2020.
\newblock \doi{10.1002/met.1961}.
\newblock URL \url{https://doi.org/10.1002%2Fmet.1961}.

\bibitem[Kasim et~al.(2021)Kasim, Watson-Parris, Deaconu, Oliver, Hatfield,
  Froula, Gregori, Jarvis, Khatiwala, Korenaga, Topp-Mugglestone, Viezzer, and
  Vinko]{Kasim_2022}
M.~F. Kasim, D.~Watson-Parris, L.~Deaconu, S.~Oliver, P.~Hatfield, D.~H.
  Froula, G.~Gregori, M.~Jarvis, S.~Khatiwala, J.~Korenaga,
  J.~Topp-Mugglestone, E.~Viezzer, and S.~M. Vinko.
\newblock Building high accuracy emulators for scientific simulations with deep
  neural architecture search.
\newblock \emph{Machine Learning: Science and Technology}, 3\penalty0
  (1):\penalty0 015013, dec 2021.
\newblock \doi{10.1088/2632-2153/ac3ffa}.
\newblock URL \url{https://dx.doi.org/10.1088/2632-2153/ac3ffa}.

\bibitem[Kovachki et~al.(2023)Kovachki, Li, Liu, Azizzadenesheli, Bhattacharya,
  Stuart, and Anandkumar]{kovachki_2023}
N.~Kovachki, Z.~Li, B.~Liu, K.~Azizzadenesheli, K.~Bhattacharya, A.~Stuart, and
  A.~Anandkumar.
\newblock Neural operator: Learning maps between function spaces, 2023.

\bibitem[Leinonen et~al.(2021)Leinonen, Nerini, and Berne]{Leinonen_2021}
J.~Leinonen, D.~Nerini, and A.~Berne.
\newblock Stochastic super-resolution for downscaling time-evolving atmospheric
  fields with a generative adversarial network.
\newblock \emph{{IEEE} Transactions on Geoscience and Remote Sensing},
  59\penalty0 (9):\penalty0 7211--7223, sep 2021.
\newblock \doi{10.1109/tgrs.2020.3032790}.
\newblock URL \url{https://doi.org/10.1109%2Ftgrs.2020.3032790}.

\bibitem[Li et~al.(2020)Li, Kovachki, Azizzadenesheli, Liu, Bhattacharya,
  Stuart, and Anandkumar]{li2020neuop}
Z.~Li, N.~Kovachki, K.~Azizzadenesheli, B.~Liu, K.~Bhattacharya, A.~Stuart, and
  A.~Anandkumar.
\newblock Neural operator: Graph kernel network for partial differential
  equations, 2020.
\newblock URL \url{https://arxiv.org/abs/2003.03485}.

\bibitem[Li et~al.(2021)Li, Kovachki, Azizzadenesheli, liu, Bhattacharya,
  Stuart, and Anandkumar]{li2021Fourier}
Z.~Li, N.~B. Kovachki, K.~Azizzadenesheli, B.~liu, K.~Bhattacharya, A.~Stuart,
  and A.~Anandkumar.
\newblock Fourier neural operator for parametric partial differential
  equations.
\newblock In \emph{International Conference on Learning Representations}, 2021.
\newblock URL \url{https://openreview.net/forum?id=c8P9NQVtmnO}.

\bibitem[Liu et~al.(2017)Liu, Yeh, Tang, Liu, and Agarwala]{Liu_2017}
Z.~Liu, R.~A. Yeh, X.~Tang, Y.~Liu, and A.~Agarwala.
\newblock Video frame synthesis using deep voxel flow.
\newblock In \emph{2017 {IEEE} International Conference on Computer Vision
  ({ICCV})}. {IEEE}, oct 2017.
\newblock \doi{10.1109/iccv.2017.478}.
\newblock URL \url{https://doi.org/10.1109%2Ficcv.2017.478}.

\bibitem[McCoy et~al.(2020)McCoy, McCoy, Wood, Regayre, Watson-Parris,
  Grosvenor, Mulcahy, Hu, Bender, Field, et~al.]{mccoy_2020}
I.~L. McCoy, D.~T. McCoy, R.~Wood, L.~Regayre, D.~Watson-Parris, D.~P.
  Grosvenor, J.~P. Mulcahy, Y.~Hu, F.~A.-M. Bender, P.~R. Field, et~al.
\newblock The hemispheric contrast in cloud microphysical properties constrains
  aerosol forcing.
\newblock \emph{Proceedings of the National Academy of Sciences}, 117\penalty0
  (32):\penalty0 18998--19006, 2020.

\bibitem[Price and Rasp(2022)]{Price_2022}
I.~Price and S.~Rasp.
\newblock Increasing the accuracy and resolution of precipitation forecasts
  using deep generative models, 2022.
\newblock URL \url{https://arxiv.org/abs/2203.12297}.

\bibitem[Serifi et~al.(2021)Serifi, Günther, and Ban]{Serifi_2021}
A.~Serifi, T.~Günther, and N.~Ban.
\newblock Spatio-temporal downscaling of climate data using convolutional and
  error-predicting neural networks.
\newblock \emph{Frontiers in Climate}, 3, apr 2021.
\newblock \doi{10.3389/fclim.2021.656479}.
\newblock URL \url{https://doi.org/10.3389%2Ffclim.2021.656479}.

\bibitem[Tran et~al.(2021)Tran, Leonarduzzi, la~Fuente, Hull, Bansal,
  Chennault, Gentine, Melchior, Condon, and Maxwell]{Tran_2021}
H.~Tran, E.~Leonarduzzi, L.~D. la~Fuente, R.~B. Hull, V.~Bansal, C.~Chennault,
  P.~Gentine, P.~Melchior, L.~E. Condon, and R.~M. Maxwell.
\newblock Development of a deep learning emulator for a distributed
  groundwater{\textendash}surface water model: {ParFlow}-{ML}.
\newblock \emph{Water}, 13\penalty0 (23):\penalty0 3393, dec 2021.
\newblock \doi{10.3390/w13233393}.
\newblock URL \url{https://doi.org/10.3390%2Fw13233393}.

\bibitem[Wang et~al.(2018)Wang, Yu, Wu, Gu, Liu, Dong, Loy, Qiao, and
  Tang]{Wang_2018}
X.~Wang, K.~Yu, S.~Wu, J.~Gu, Y.~Liu, C.~Dong, C.~C. Loy, Y.~Qiao, and X.~Tang.
\newblock Esrgan: Enhanced super-resolution generative adversarial networks,
  2018.
\newblock URL \url{https://arxiv.org/abs/1809.00219}.

\bibitem[Watson et~al.(2020)Watson, Wang, Lynar, and Weldemariam]{Watson_2020}
C.~D. Watson, C.~Wang, T.~Lynar, and K.~Weldemariam.
\newblock Investigating two super-resolution methods for downscaling
  precipitation: Esrgan and car, 2020.
\newblock URL \url{https://arxiv.org/abs/2012.01233}.

\bibitem[Watson-Parris(2021)]{Watson_Parris_2021}
D.~Watson-Parris.
\newblock Machine learning for weather and climate are worlds apart.
\newblock \emph{Philosophical Transactions of the Royal Society A:
  Mathematical, Physical and Engineering Sciences}, 379\penalty0
  (2194):\penalty0 20200098, feb 2021.
\newblock \doi{10.1098/rsta.2020.0098}.
\newblock URL \url{https://doi.org/10.1098%2Frsta.2020.0098}.

\end{thebibliography}

\end{document}